# 基于 YOLOv8n 的轻量化虾病害检测研究

费玉环，王庚辰，刘丰豪，臧 冉，孙旭菲，常 昊

（曲阜师范大学工学院，276826，山东省日照市）

**摘要**：虾病害是导致虾养殖业经济损失的主要原因之一，为了提升虾养殖领域的智能检测效率，本文提出了一种基于 YOLOv8n 的轻量化网络架构。首先通过设计 RLDD 检测头和 C2f-EMCM 模块，在保证检测精度的同时减少了模型的参数量和计算量，提高了计算效率。随后，引入改进的 SegNext_Attention 自注意力机制，进一步增强模型的特征提取能力，使其能够更准确地捕捉虾病特征。最终，在自建虾病数据集上进行消融实验和对比实验，并扩展到 URPC2020 数据集上进行泛化实验。结果表明，所提出的模型参数量比原有的YOLOv8n减少32.3%，mAP@0.5 值为92.7%，比YOLOv8n 增加3%；同时提出的模型在mAP@0.5、参数量、模型尺寸方面均优于其他 YOLO 系列的轻量化模型。综上所述，所提出的方法在精度和效率之间取得了良好平衡，为虾养殖领域的疾病智能检测提供了可靠的技术支持。

## 1　引言

虾病害是虾养殖业面临的主要挑战之一，对虾农的经济造成严重影响[1,2]。传统虾病诊断方法主要包括视觉检测法和实验室病原体分离鉴定法。前者主观性强、准确率低且人工投入大，后者操作复杂、耗时长、成本高，难以满足实时检测需求。近年来，基于深度学习的目标检测算法在计算机视觉领域得到广泛应用，其中 YOLO（You Only Look Once）[3]系列检测算法凭借其实时性能与较高的检测精度，成为研究热点。YOLO 系列通过将目标检测任务转化为回归问题，实现了端到端的检测流程，保持较高推理速度。然而，YOLO 系列检测算法在微小特征检测方面仍受特征提取能力限制，导致漏检率较高。针对上述问题，研究者们从网络架构优化和特征增强两个维度提出了多种创新性改进方案。

在网络架构优化方面，辛世澳等[4]采用轻量级 ShuffleNetv2 结合双向特征金字塔网络（BIFPN）对 YOLOv7 进行优化，显著提升了特征提取效率并降低了计算复杂度；刘向举等[5]提出的 DCN-YOLOv5 算法通过设计可变形卷积空间融合模块，有效提升了目标边界模糊场景下的检测精度；Wen 等[6]将选择性内核网络与 YOLOv5 的主干网络相融合，通过加权不同尺度通道的特征信息，有效增强了模型对模糊目标的识别能力。在特征增强方面，Xia 等[7]在 YOLOv4 中引入感受野块结构，通过扩大特征感受野增强了小目标的细节和位置信息，使平均精度提升了 5.1%；翟先一等[8]针对水下环境对比度低的问题，提出了一种基于改进 YOLOv5s 的检测方法，通过多尺度视觉恢复算法增强图像对比度，显著提升了模型的特征提取能力；Hou 等[9]在 YOLOv5s 中引入 gnConv 自注意力机制，增强了模型对多尺度特征的识别能力；Li 等[10]通过三重注意力机制优化 YOLOv5 的 Neck 结构，显著提升了小目标检测性能；Yu 等[11]基于 YOLOv7 模型创新性地将 CrossConv 卷积与 3D 注意力机制相结合，有效增强了模型在水下复杂环境中的抗干扰能力；Liang 等[12]通过将多信息流融合注意力机制 SGCA 嵌入 YOLOv7 框架，显著改善了低可见度环境下的检测效果。

然而，面对形态特征细微、目标遮挡和背景干扰等挑战，现有方法仍存在改进空间。为了解决这些问题，本文提出了一种基于

收稿日期:
基金项目:国家自然科学基金(51705286)；山东省自然科学基金(ZR2017BEE057)；山东省自然科学基金(ZR2019MEE054)；山东省高等学校“青创科技计划”项目(2019KJN036).
通信作者:费玉环, 女, 1982-, 博士, 副教授；研究方向：计算机模拟，机器学习；E-mail: yuhuanfei@qfnu.edu.cn.

YOLOv8n 的轻量化且更高效的目标检测模型，本文的主要工作如下：

（1）提出了全新的重参数轻量化检测头（Reparameterized Lightweight Disease Detection, RLDD）。该设计采用重参数化技术，在训练阶段通过多分支结构增强特征提取能力，在推理阶段等价转换为单一卷积层，实现了无损的模型加速。

（2）设计了高效多尺度卷积模块（Efficient Multi-scale Convolution Module, EMCM），替换 Bottleneck 模块中的标准卷积。使模块能够在减少计算量的同时，增强模型对不同尺度目标的区分能力，特别适用于目标尺度变化较大的场景。

（3）引入改进的 SegNext_Attention 自注意力机制，通过并行空间和通道注意力模块，学习像素级的注意力权重，使模型能够自适应地聚焦于关键区域，同时有效抑制制背景干扰，增强了模型的特征提取能力。

## 2　改进的 YOLOv8 模型

### 2.1　RLDD 重参数轻量化检测头

YOLOv8n 的检测头采用了解耦头设计，使分类和回归任务相互独立，其结构如图 1(a)所示。这种设计能够减少任务间的相互干扰，增强特征提取的专用性，从而提高检测精度和训练速度。然而，每个输出层级都需要分别通过两个 3×3 卷积层和一个 1×1 卷积层来提取定位和分类信息，增加了模型的参数量和计算复杂度。此外，不同层级特征图进行独立预测，缺乏层级间的特征信息共享和融合，在一定程度上限制了模型在多尺度目标检测中的性能。

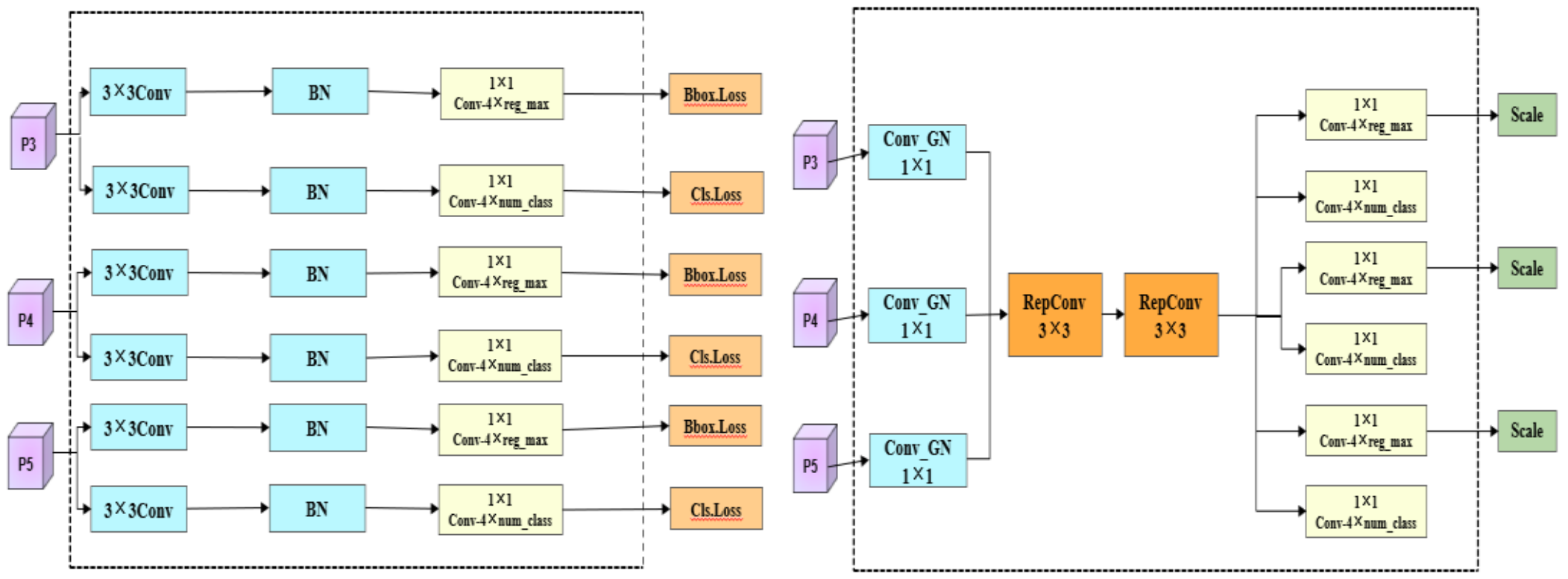

(a) YOLOv8n 检测头的网络结构　　(b) RLDD 检测头的网络结构

图 1 YOLOv8n 检测头和 RLDD 检测头网络结构图

为了解决上述问题，本文提出了一种轻量级的重参数化检测头 RLDD，其结构如图 1(b)所示。RLDD 在保留解耦头双分支结构的基础上，通过引入重参数化卷积使其在训练阶段通过 3×3 卷积、1×1 卷积和 3×3 平均池化等多分支结构，增强模型对多尺度特征的捕捉能力。在推理阶段，通过结构重参数化技术将多分支结构等价转换为单一 3×3 卷积层，减少参数量。

具体而言，RLDD 接收 P3、P4、P5 输入特征图后，首先通过 1×1 卷积进行初步的特征提取和通道统一，以确保不同层级特征图的通道数一致。随后，每个层级的特征图依次经过两个 3×3 的重参数化卷积 RepConv，使得模型在捕捉更丰富特征信息的同时减少了参数量，实现无损加速。最后，经过边界框预测卷积生成包含边界框回归结果的特征图，并通过缩放因子进行尺度变换，以确保这些回归结果适应不同特征层的分辨率。同时，经过分类预测卷积生成包含分类结果的特征图。实验结果证明该模块的引进弥补了模型轻量化后可能带来的精度丢失的问题，提高了检测效率。

### 2.2　C2f-EMCM

在 YOLOv8n 的 C2f 模块中，尽管通过跨阶段部分连接和深度可分离卷积实现了特征复用，但其设计仍存在不足。首先，C2f 模块默认采用固定尺寸的 3×3 卷积核进行

特征提取，难以提取虾病中不同尺度的特征。其次，虽然 C2f 模块通过多分支 Bottleneck 结构增强特征多样性，但重复的 3×3 卷积操作仍引入冗余计算。

针对 C2f 模块的不足，本文提出一种基于多尺度特征融合的卷积 EMCM，来替换 Bottleneck 模块中的标准卷积，得到全新的 C2f-EMCM 模块。EMCM 卷积的结构如图 2 所示，其核心是采用通道分组策略实现特征的多路径处理。首先，将输入特征沿通道维度分割为两组，其中一组通过特征复用策略划分为直接保留原始信息的"原始特征"，另一组则对原始信息进行拆分，分别采用 3×3 和 5×5 两种不同尺寸的卷积核进行并行特征提取，来捕捉多尺度特征信息。随后，将提取的多尺度特征与原始特征进行通道拼接，并引入 1×1 卷积实现跨通道交互与维度调整，使该模块在保证多尺度信息完整性的同时，降低参数冗余，使模型更容易部署到资源受限的边缘设备上。

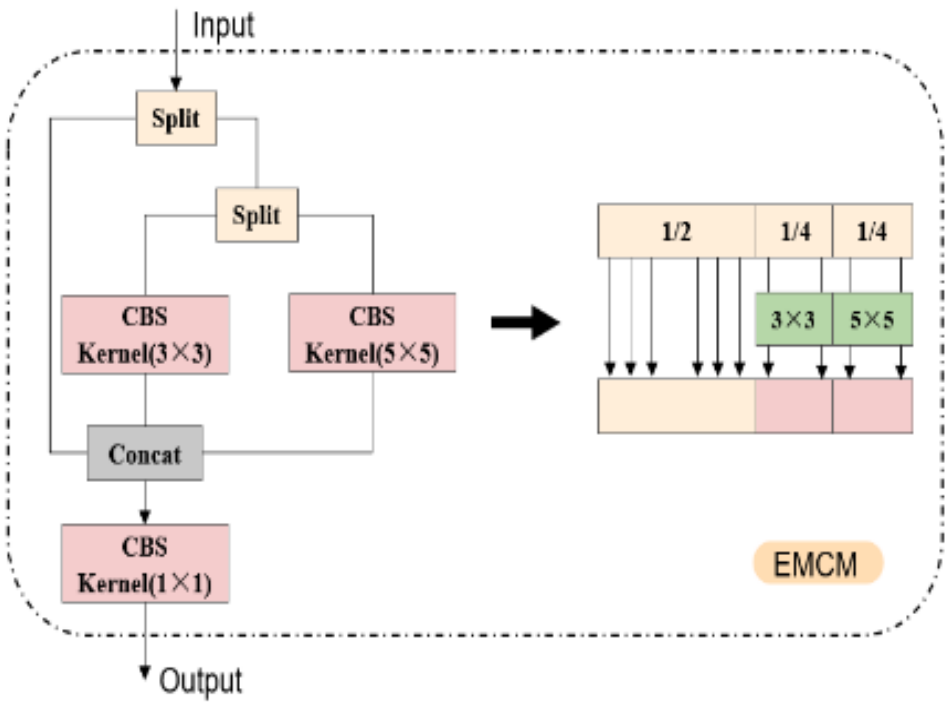

图 2 EMCM 模块结构图

将提出的 EMCM 卷积替换 Bottleneck 模块中的标准卷积，构成新的 Bottleneck-EMCM 模块，如图 3(a)所示。然后将 C2f 中的 Bottleneck 模块替换成新建的 Bottleneck-EMCM 模块，从而形成新的 C2f-EMCM 模块，如图 3(b)所示。最终，将 YOLOv8n 主干网络中的最后两个C2f模块以及颈部网络中的第一个和最后两个C2f模块替换为能够同时提取细节纹理特征和形状特征的 C2f-EMCM 模块。经过上述改进，一方面通过多尺度特征融合提升了微小病害的检测能力，另一方面通过引入跨尺度融合机制，有效缓解了大面积病变因背景干扰导致检测性能较差的问题。

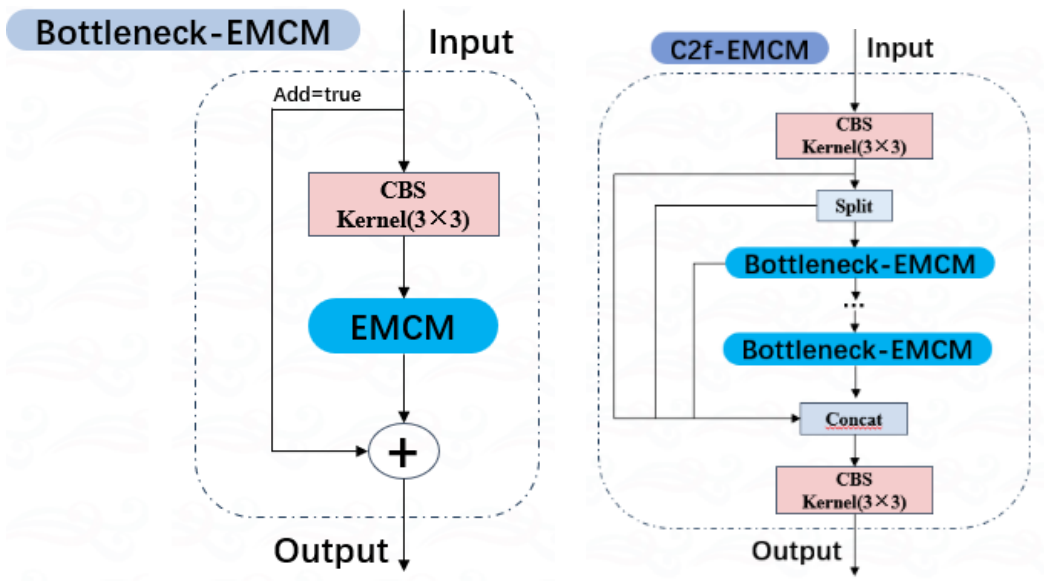

(a) Bottleneck-EMCM (b) C2f-EMCM

图 3 Bottleneck-EMCM 模块和 C2f-EMCM 模块结构图

## 2.3 SegNext_Attention 自注意力模块

SegNext_Attention 通过引入基于分解的 Hamburger 模块[13]来提取全局上下文信息，同时采用大卷积核注意力机制以保持与大规模感受野特征的一致性。多尺度卷积注意力模块（Multi-Scale Convolutional Attention, MSCA）是 SegNext_Attention 的核心组件，结构如图 4 所示。MSCA 模块通过深度卷积实现局部信息聚合，并利用深度可分离卷积捕获多尺度上下文信息，这种设计在保持多尺度特征提取能力的同时，显著降低了计算复杂度，其数学形式如公式(1)，公式(2)所示。

$$Att=Conv_{1\times1}(\sum_{i=0}^{3}Scale_i(DW.Conv\text{-}Conv(f))) \quad (1)$$

$$Out=Att\otimes f \quad (2)$$

其中，DW.Conv表示深度可分离卷积操作，$Scale_i$表示不同尺度的特征缩放，$\otimes$表示逐元素矩阵乘法运算，f 表示输入特征，Out 和 Att 分别表示输出和注意力图。

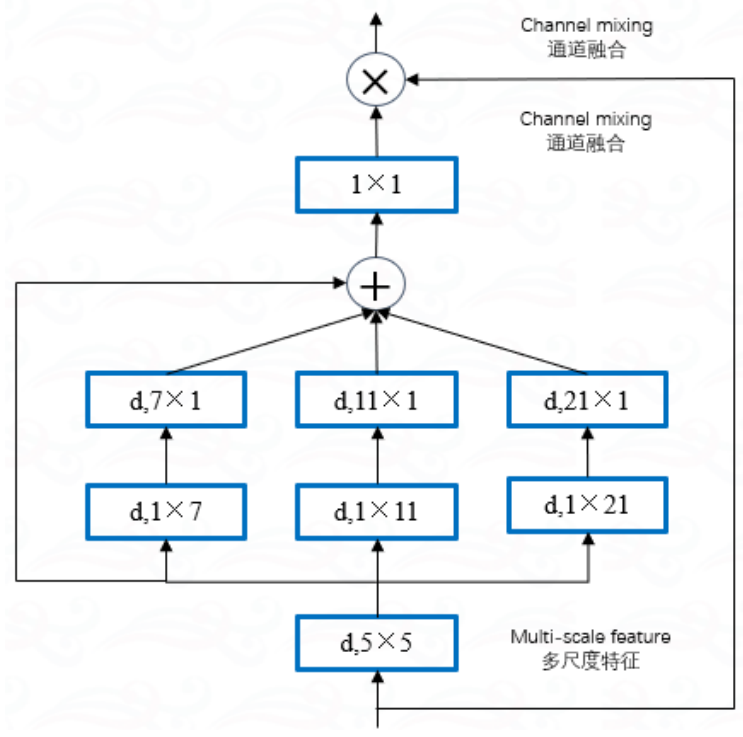

图 4 MSCA 模块结构图

SegNext_Attention 通过引入轻量级解码器优化了计算效率，省略了传统特征提取的初始阶段，避免了过多低级特征带来的计

算冗余和性能瓶颈。同时，为进一步增强模型的全局建模能力，SegNext_Attention 在编码器与解码器之间嵌入了注意力机制，使其能够自适应地分配像素级注意力权重。将 SegNext_Attention 引入到 YOLOv8n 的骨干网络后，模型能够更精确地聚焦于目标区域，同时有效抑制背景干扰。

## 3 实验设计与结果分析

### 3.1 实验设计

#### 3.1.1 实验参数与实验环境

本文训练模型所用的实验环境的操作系统为 Windows 11、编程语言为 Python 3.9、CPU 为 Intel(R) Core(TM) i7-13650HX、GPU 为 GeForce RTX 4060，算法框架为 PyTorch 1.19.0、GPU 加速库为 CUDA 11.1、CUDNN 8.6.0。主要训练参数如表 1 所示。

表 1 训练参数配置

| 参数 | 配置 |
|---|---|
| 输入图像尺寸 | 640×640 |
| 初始学习率 | 0.01 |
| 动量系数 | 0.937 |
| 权重衰减系数 | 0.0005 |
| 每批次的样本数 | 16 |
| 迭代次数 | 300 |

#### 3.1.2 数据集与预处理

本实验使用的虾病数据集主要从互联网自行收集、整理并进行标注，共包含 1047 张图像，涵盖三种典型的虾病类别：白斑病（WSSV）、黑鳃病（BSS）和黄鳃病（SBGS）。为确保实验的可靠性和泛化能力，数据集按 9:1 的比例随机划分为训练集和验证集，并将输入图像的尺寸统一处理为 640×640 像素，部分数据集图像如图 5 所示。

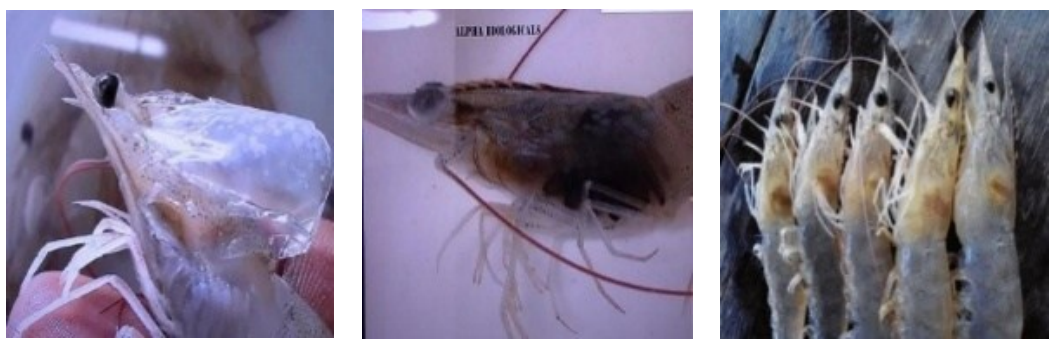

(a) 白斑病 (b) 黑鳃病 (c) 黄鳃病

图 5 虾病害检测数据集

#### 3.1.3 模型评价指标

在虾病害检测中，从模型对病害的识别与定位能力角度考虑，选取精确度(Precision, P)、召回率(Recall, R)、平均精度均值(Mean Average Precision，mAP@0.5)作为评价指标，具体计算公式为：

$$P=\frac{TP}{TP+FP} \tag{3}$$

$$R=\frac{TP}{TP+FN} \tag{4}$$

$$mAP@0.5=\frac{1}{C}\sum_{K=1}^{C}AP@0.5 \tag{5}$$

$$AP@0.5=\frac{1}{n}P_1+\frac{1}{n}P_2+...+\frac{1}{n}P_n \tag{6}$$

其中，TP 表示正确检测的目标数量，FP 表示错误检测的目标数量，FN 表示未检测到的实际目标数量，n 表示类别数。

从模型的轻量化程度与计算资源需求角度考虑，选取参数量（Parameters）作为评价指标。

### 3.2 实验结果分析

#### 3.2.1 消融实验

为了验证各项改进的有效性，本文在同一实验环境下进行了消融实验。选择 YOLOv8n 作为基准模型，分别添加 RLDD、EMCM、SegNext_Attention 进行定量分析，共设计 8 组实验，消融实验结果如表 3 所示。

从表 3 中的消融实验结果可以看出，相较于实验 1(直接使用 YOLOv8n 作为基准模型)的 78.4%的精确度、90.2%的召回率和 89.7%的 mAP@0.5，实验 2(单独采用 RLDD 检测头)将精确度提升 5.3%，参数量从 3.1M 减少至 2.3M，降幅达 25.8%。实验 3(单独采用 EMCM 模块)表现最为突出，其单独使用时将精确度提升至 86.5%，较基准模型提升 8.1%，同时 mAP@0.5 提升 2.4%至 92.1%。实验 4(单独采用 SegNext_Attention 注意力模块)的精确度以及 mAP@0.5 较基准模型均有小幅提升。双模块组合中，实验 6(融合 RLDD 检测头以及 SegNext_Attention 注意力模块)以 2.5M 的参数量实现了 86.9%的精确度、90.0%的召回率和 92.1%的 mAP@0.5，综合性能最优。而实验 7(融合 EMCM 模块以及 SegNext_Attention 注意力模块)则导致

精确度回退至基准水平 78.4%，显示出仅使用这两模块效果一般。最终实验 8(融合所有改进模块)以 2.1M 的参数量实现综合提升，精确度达 87.5%，较基准模型提升 9.1%，mAP@0.5 提升 3.0%至 92.7%，其精确度和 mAP@0.5 分别较单模块最高值提升 1%和 0.6%，以及双模块组合的最高值提升 0.6%和 0.6%，证明了最终改进模块在检测精度和轻量化方面改进的有效性。

综上所述，RLDD 检测头减少 25.8%参数量并使精确度提升 5.3%，EMCM 模块使精确度提升 8.1%，成为核心特征增强单元，而 SegNext_Attention 模块单独作用有限，但与 RLDD 检测头协同作用可达成最优双模块组合。全模块融合后，模型以最低参数量实现最高精确度和 mAP@0.5，较基准模型分别提升 9.1%和 3.0%，验证了最终改进模型有效平衡了检测精度与轻量化需求，为复杂场景虾病害检测提供了解决方案。

表 3 消融实验设计及结果

| 模型 | RLDD | EMCM | SegNext_Attention | 精确度(%) | 召回率(%) | mAP@0.5(%) | 参数量(M) |
|---|---|---|---|---|---|---|---|
| 实验 1 | - | - | - | 78.4 | 90.2 | 89.7 | 3.1 |
| 实验 2 | √ | - | - | 83.7 | 87.1 | 91.5 | 2.3 |
| 实验 3 | - | √ | - | 86.5 | 87.7 | 92.1 | 2.7 |
| 实验 4 | - | - | √ | 82.0 | 87.2 | 90.5 | 3.1 |
| 实验 5 | √ | √ | - | 86.3 | 84.7 | 91.5 | 2.1 |
| 实验 6 | √ | - | √ | 86.9 | 90.0 | 92.1 | 2.5 |
| 实验 7 | - | √ | √ | 78.4 | 91.3 | 89.7 | 3.0 |
| 实验 8 | √ | √ | √ | 87.5 | 88.9 | 92.7 | 2.1 |

### 3.2.2 对比实验

为了进一步验证本文所改进模型的性能，选取当前主流的目标检测算法 YOLOv5[14]、YOLOv8n[15]、YOLOv10[16]、YOLOv11[17]、FasterNet[18]、BiFPN[19]做对照，所有实验均在相同的平台和数据集条件下进行，对比实验结果如表 4 所示。

表 4 对比实验结果

| 模型 | 精确度(%) | 召回率(%) | mAP@0.5(%) | 参数量(M) |
|---|---|---|---|---|
| YOLOv5 | 87.3 | 85.3 | 87.4 | 2.5 |
| YOLOv8n | 78.4 | 90.2 | 89.7 | 3.1 |
| YOLOv10 | 86.9 | 79.7 | 86.9 | 2.7 |
| YOLOv11 | 84.6 | 87.7 | 88.8 | 2.6 |
| FasterNet | 86.8 | 77.4 | 87.8 | 4.2 |
| BiFPN | 83.8 | 87.0 | 88.1 | 2.0 |
| 改进模型 | 87.5 | 88.9 | 92.7 | 2.1 |

根据表 4 的结果可以看出，本文改进的模型在虾病害检测任务中表现出显著的优势，其精确度达到 87.5%，较 YOLOv8n 的 78.4%提升 9.1%，同时优于 BiFPN 的 83.8%和 YOLOv11 的 84.6%，表明改进模型在正样本识别上具有更高的准确性，能够有效降低误检率。在召回率方面，改进模型达到 88.9%，相比于 YOLOv5、YOLOv10、

YOLOv11、FasterNet 和 BiFPN 提升了 3.6%、9.2%、1.2%、11.5%、1.9%，说明其在捕捉正样本方面表现更为全面，减少了漏检的可能性。此外，改进模型的 mAP@0.5 为 92.7%，超过所有对比模型，进一步验证了其在多尺度目标检测和复杂背景下的鲁棒性与稳定性。在模型效率方面，改进模型在参数量和计算复杂度上进行了优化，参数量较 YOLOv5、YOLOv8n、YOLOv10、YOLOv11、FasterNet 分别降低了 16%、32.3%、22.2%、19.2%、50%，仅较 BiFPN 略有增加，但综合考虑模型的整体性能，该差值可忽略。

综合来看，改进模型通过结构优化和特征提取策略的改进，提升了虾病害检测的精度，同时保持了较高的计算效率，为虾病害检测提供了一个高效的解决方案。

### 3.2.3 可视化分析

为了更加直观地展示改进后模型的检测性能，本文以部分典型测试集图像为基准，选取了 YOLOv8n 与改进后模型进行可视化对比实验，可视化结果如图 6 所示。

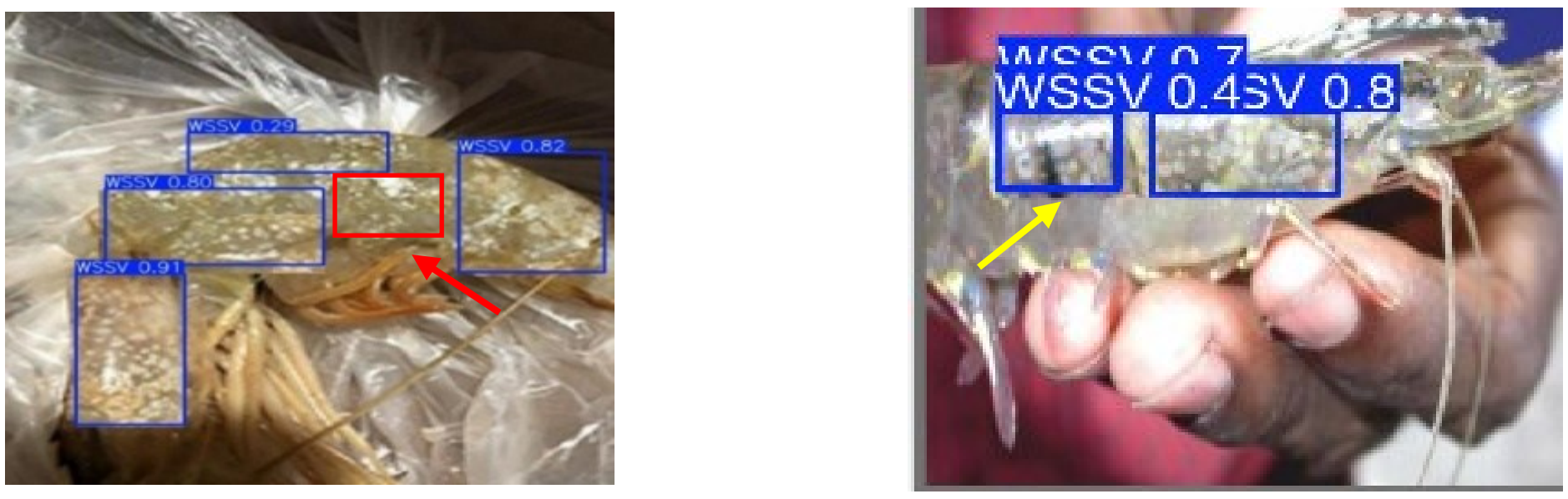

(a) YOLOv8n 可视化结果图

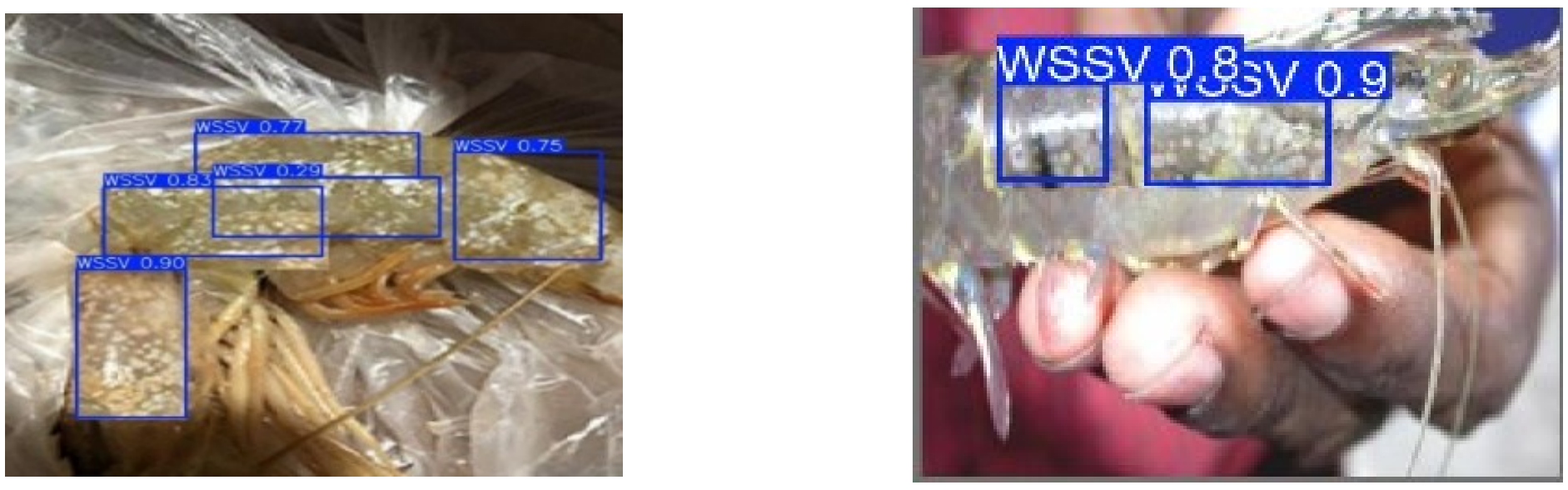

(b) 改进模型可视化结果图

图 6 YOLOv8n 及改进模型可视化结果图

从可视化结果图来看，改进后的模型在整体性能上表现出优势，其相较于 YOLOv8n 展现出更强的特征提取能力和抗干扰能力，能够更准确地识别目标，此外，在复杂背景或目标特征不显著的情况下，YOLOv8n 容易出现漏检(如图 6(a)中红色箭头所示)且检测框之间存在重叠(如图 6(a)中黄色箭头所示)的现象。相比之下，改进模型通过引入自注意力机制以及多尺度特征融合机制等，提升了目标特征的区分度，有效缓解了检测框重叠的问题，并降低了漏检率。

综上所述，改进后的模型可以有效识别复杂背景下的虾病害，能够减少漏检与检测框重叠的发生。这表明改进后的模型对特征的提取能力有所提升，能够更好的捕捉图像中的细节信息，有效抑制噪声干扰，展现出良好的检测性能，为实际虾病害检测提供了有力支撑。

### 3.2.4 泛化实验

为了进一步验证模型在实际应用中的通用性和泛化能力，本文选取 URPC2020 数据集进行泛化实验，该数据集来自 2020 年全国水下机器人大赛，主要用于目标检测、分割和跟踪等计算机视觉任务，数据集中包括海星(starfish)、海胆(echinus)、海参(holothurian)等多个类别，泛化实验结果如表 5 所示。

表 5 泛化实验结果

| 模型 | 精确度(%) | 召回率(%) | mAP@0.5(%) | 参数量(M) |
|---|---|---|---|---|
| YOLOv8n | 82.4 | 75.9 | 83.1 | 3.1 |
| 改进模型 | 86.3 | 74.1 | 87.2 | 2.1 |

泛化实验结果表明，改进模型在 URPC2020 数据集上展现出较强的泛化能力，其精确度和 mAP@0.5 均高于 YOLOv8n，分别提升了 3.9%和 4.1%，验证了模型在不同数据分布下的适应性和抗干扰能力。并且改进模型在参数量上较 YOLOv8n 减少了 32.3%，使其在实际应用中更具优势。综合来看，改进模型在泛化能力、精度和效率之间取得了良好的平衡，为实际应用场景提供了更可靠的解决方案，泛化实验可视化结果如图 7 所示。

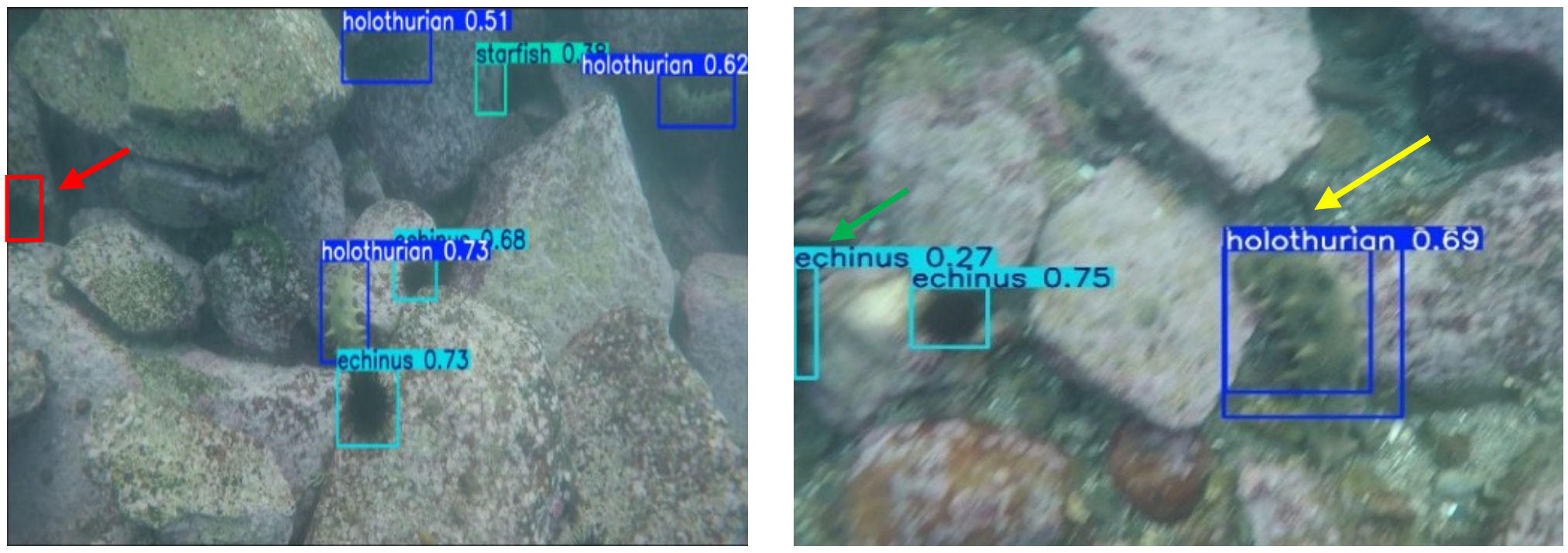

(a) YOLOv8n 泛化实验可视化结果

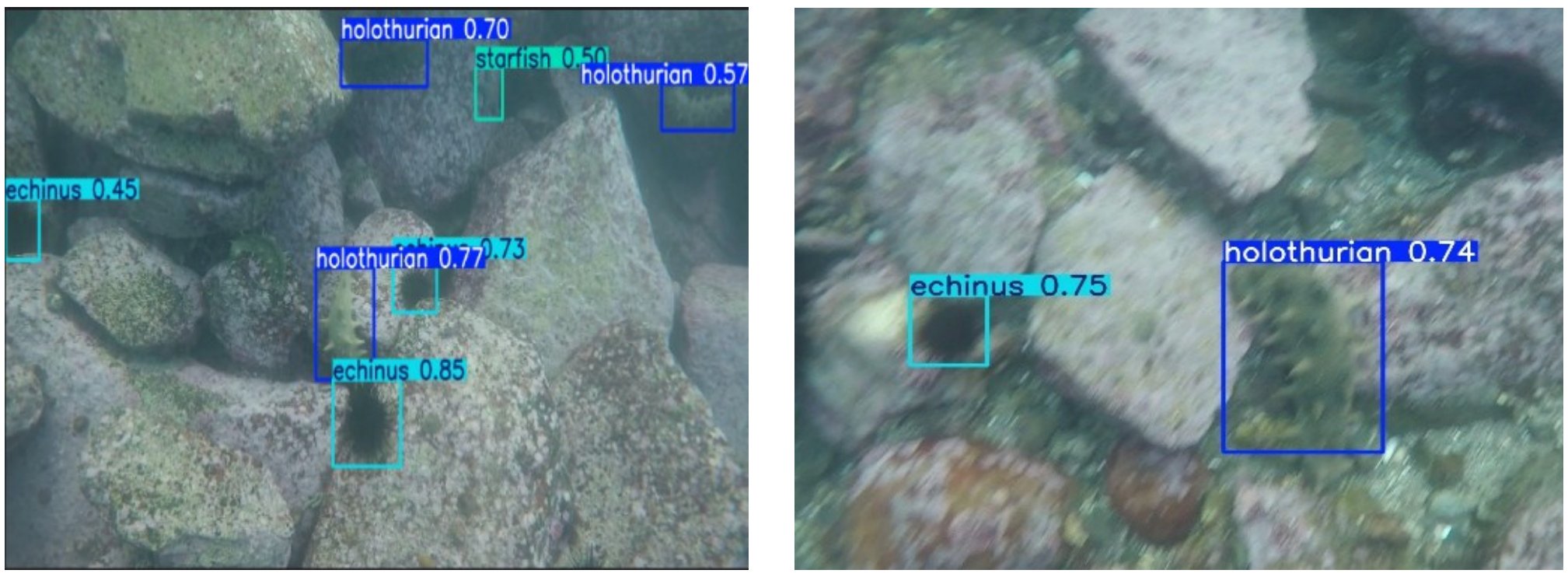

(b) 改进模型泛化实验可视化结果

图 7 YOLOv8n 及改进模型泛化实验可视化结果图

从图 7 的可视化结果可以看出，改进模型相较于 YOLOv8n 在特定类别的检测精度上提升(如海参的检测)，尤其是在目标特征模糊的场景下，改进模型能够更准确地识别目标，有效减少了漏检(如图 7(a)中红色箭头所示)以及误检(如图 7(a)中绿色箭头所示)的情况。此外，改进模型通过引入多尺度特征融合机制，降低了检测框重叠(如图 7(a)中黄色箭头所示)的问题，提升了目标定位的准确性。这些结果再次表明，改进模型具有很强的泛化能力，能够更好地适应复杂场景下的检测任务。

## 4 结论

本研究针对虾类病害检测任务，提出了一种基于 YOLOv8n 的轻量化网络架构，通过提出 RLDD 可重参数化检测头、EMCM 高效多尺度卷积模块且引入了改进的 SegNext_Attention 自注意力机制，提升了模型的检测精度和计算效率。对比实验结果表明，改进模型在精确度、召回率和 mAP@0.5 等关键指标上优于 YOLOv5、YOLOv8n、

YOLOv10、YOLOv11、FasterNet 和 BiFPN 等主流模型。较基准模型 YOLOv8n，mAP@0.5 提高了 3%，达到了 92.7%。此外，改进模型的召回率和精确度高于其他对比模型，验证了其在复杂背景下的抗干扰能力和稳定性。同时改进模型在参数量上较 YOLOv8n 减少 32.3%，使其更适合在资源受限的边缘设备上部署。最后，在 URPC2020 数据集上的泛化实验进一步验证了其泛化能力，mAP@0.5 较 YOLOv8n 提升了 4.1%。

综上所述，本研究通过结构优化和特征提取策略的改进，成功实现了虾病害检测的高精度与轻量化的目标，为虾养殖领域的病害智能检测提供了新的技术思路。未来的研究将进一步探索模型在更多病害类型和复杂环境下的应用，以进一步提升其泛化能力和实际应用价值。

## 参考文献：

[1] YU Y B, CHOI J H, KANG J C, et al. Shrimp bacterial and parasitic disease listed in the OIE: A review[J]. Microbial Pathogenesis, 2022, 166: 105545.

[2] LEE D, YU Y B, CHOI J H, et al. Viral shrimp diseases listed by the OIE: A review[J]. Viruses, 2022, 14(3): 585.

[3] REDMON J, DIVVALA S, GIRSHICK R, et al. You only look once: Unified, real-time object detection[C]//Proceedings of 2016 IEEE Conference on Computer Vision and Pattern Recognition. 2016: 779-788. DOI: 10.1109/CVPR.2016.91.

[4] 辛世澳, 葛海波, 袁昊, 等. 改进 YOLOv7 的轻量化水下目标检测算法[J]. 计算机工程与应用, 2023, 60(3): 88-99.

[5] 刘向举, 刘洋, 蒋社想. 基于 SimAM 注意力机制的 DCN-YOLOv5 水下目标检测[J]. 重庆工商大学学报(自然科学版), 2023, 40(2): 76-81.

[6] WEN G, LI S, LIU F, et al. YOLOv5s-CA: A modified YOLOv5s network with coordinate attention for underwater target detection[J]. Sensors, 2023, 23(7): 3367.

[7] XIA C, FU L, LIU H, et al. In situ sea cucumber detection based on deep learning approach[C]//2018 OCEANS MTS/IEEE Kobe Techno-Oceans (OTO). Kobe, Japan: IEEE, 2018: 1-4.

[8] 翟先一, 魏鸿磊, 韩美奇, 等. 基于改进 YOLO 卷积神经网络的水下海参检测[J]. 江苏农业学报, 2023, 39(7): 1543-1553.

[9] HOU C, GUAN Z, GUO Z, et al. An improved YOLOv5s-based scheme for target detection in a complex underwater environment[J]. Journal of Marine Science and Engineering, 2023, 11(5): 1041.

[10] LI Y, BAI X, XIA C. An improved YOLOV5 based on triplet attention and prediction head optimization for marine organism detection on underwater mobile platforms[J]. Journal of Marine Science and Engineering, 2022, 10(9): 1230.

[11] YU G, CAI R, SU J, et al. U-YOLOv7: A network for underwater organism detection[J]. Ecological Informatics, 2023, 75: 102108.

[12] LIANG X M, RAN L I, HAI Y U, et al. Improved Underwater Object Detection Algorithm of YOLOv7[J]. Journal of Computer Engineering & Applications, 2024, 60(6).

[13] GENG Z, GUO M H, CHEN H, et al. Is attention better than matrix decomposition?[J]. arXiv preprint arXiv:2109.04553, 2021.

[14] WANG Z, JIN L, WANG S, et al. Apple stem/calyx real-time recognition using YOLO-v5 algorithm for fruit automatic loading system[J]. Postharvest Biology and Technology, 2022, 185: 111808.

[15] HUANG Z, LI L, KRIZECK G C, et al. Research on traffic sign detection based on improved YOLOv8[J]. Journal of Computer and Communications, 2023, 11(7): 226-232.

[16] WANG C Y, BOCHKOVSKIY A, LIAO H Y M. YOLOv10: Real-time end-to-end object detection[J/OL]. arXiv preprint arXiv:2305.10530, 2023.

[17] LI X, WANG W, HU X. YOLOv11: Dynamic feature fusion for real-time detection[J]. IEEE Transactions on Pattern Analysis and Machine Intelligence, 2024, 46(3): 1025-1038.

[18] CHEN J, KAO S H, HE H, et al. Run, don’t walk: Chasing higher FLOPS for faster neural n

etworks[C]//Proceedings of 2023 IEEE/CVF Conference on Computer Vision and Pattern Recognition (CVPR). Vancouver, Canada, 2023: 12021-12031.

[19] TAN M, PANG R, LE Q V. EfficientDet: Scalable and efficient object detection[C]//Proceedings of 2020 IEEE/CVF Conference on Computer Vision and Pattern Recognition (CVPR). 2020: 10781-10790.

# Research on Lightweight Shrimp Disease Detection Based on YOLOv8n

*FEI Yuhuan, WANG Gengchen, LIU Fenghao, ZANG Ran, SUN Xufei, CHANG Hao*

(College of Engineering, Qufu Normal University, 276826, Rizhao, Shandong, PRC)

**Abstract:** Shrimp diseases are one of the primary causes of economic losses in shrimp aquaculture. To prevent disease transmission and enhance intelligent detection efficiency in shrimp farming, this paper proposes a lightweight network architecture based on YOLOv8n. First, by designing the RLDD detection head and C2f-EMCM module, the model reduces computational complexity while maintaining detection accuracy, improving computational efficiency. Subsequently, an improved SegNext_Attention self-attention mechanism is introduced to further enhance the model's feature extraction capability, enabling more precise identification of disease characteristics. Extensive experiments, including ablation studies and comparative evaluations, are conducted on a self-constructed shrimp disease dataset, with generalization tests extended to the URPC2020 dataset. Results demonstrate that the proposed model achieves a 32.3% reduction in parameters compared to the original YOLOv8n, with a mAP@0.5 of 92.7% (3% improvement over YOLOv8n). Additionally, the model outperforms other lightweight YOLO-series models in mAP@0.5, parameter count, and model size. Generalization experiments on the URPC2020 dataset further validate the model's robustness, showing a 4.1% increase in mAP@0.5 compared to YOLOv8n. The proposed method achieves an optimal balance between accuracy and efficiency, providing reliable technical support for intelligent disease detection in shrimp aquaculture.